# A hybrid pipeline for dynamic ontology-based semantic mapping

**Konstantinos Dimitropoulos**
Computer Engineering and Informatics Department
University of Patras, Greece
kdimitrop@upatras.gr

**Ioannis Hatzilygeroudis**
Computer Engineering and Informatics Department
University of Patras, Greece
ihatz@ceid.upatras.gr
ORCID: 000-0002-4036-9369

## Abstract

Semantic mapping plays a crucial role in the ability of a robot to interact with objects, operate and navigate a complex environment. The most common pipeline for semantic mapping consists of geometric mapping and localization (SLAM), perception, semantic fusion and semantic representation. However, more recent works also integrate a form of prior knowledge in their application, most notably knowledge graphs or semantic scene graphs, to improve contextual understanding of the environment. In this paper, we present a hybrid pipeline for semantic mapping. Our system incorporates an external calibrated camera using homography projection for geometric mapping and localization, combined with object detection, persistent object tracking and ontology driven semantic updates to build a dynamic semantic world model. Linear regression models are also used for correction of the estimated values of real world coordinates. The system continuously updates object instances, spatial properties and semantic relations based on real time sensory data. Ontologies are selected as form of knowledge representation due to their hierarchical structure, semantic expressiveness and support for dynamic world modelling.

## 1. Introduction

The tasks of contemporary robots in a working environment have become quite complex. So, a robot's ability to understand the environment it operates in and be aware of its location at any time, becomes very important. In this regard, a map of the environment is required. In most approaches this is obtained via Simultaneous Localization And Mapping (SLAM) method [1], [2], [3], [4], [5]. With SLAM method a robot can build a map of the environment from sensory data, with sensors mounted on the robot, and simultaneously figure out its own position inside it. However, SLAM is often not adequate, because dynamic and complex environments require a deeper understanding of the objects, the entities and their relations. Therefore, perception, more often in the form of a machine learning model or segmentation networks, helps the robot detect and recognize the items and entities inside the environment. In addition, with semantic fusion these detections are combined with the map created by SLAM and the robot pose, so that a consistent semantic representation of the environment is created. Thus, the robot can build a better and more thorough understanding of the environment, its entities and their relations. This is the typical semantic mapping pipeline [6], [7], [8], [9], [10].

In recent research works, prior knowledge about the environment and its entities is often presented as an additional layer that enhances semantic mapping [11], [12], [13], [14], [15]. The integration of prior knowledge can provide the robot with necessary semantic information about the environment that can be utilized during exploration and reason about it. The robot can use this stored knowledge about object classification and object relations to interpret sensory data more intelligently. Prior knowledge is usually represented using probabilistic models to model how likely objects or scene configurations are, knowledge graphs or scene semantic graphs, where nodes represent objects and edges represent relations.

However, the semantic pipeline can be extended even further with a hybrid approach. First, while some semantic mapping systems include prior knowledge, the reasoning mechanism is lacking. Existing systems focus more on object localization and semantic labeling and reasoning is either absent or relatively limited. In addition, in many approaches the knowledge base is static and the position of objects inside the environment is often fixed; it is not dynamically being updated with data from sensory information. So, there is a need for an improvement, mostly as far as prior knowledge and reasoning are concerned, with a more structural knowledge representation that enables higher level reasoning, as well as a dynamic updating of that knowledge base.

To achieve the above objectives, in this paper, we present an approach based on sensory data from an external calibrated camera using homography projection, combined with object detection and object tracking, an ontology as prior knowledge and online real time updates of this ontology, to build a dynamic semantic map. The main contributions of this work are summarized as follows:

- A hybrid pipeline for dynamic semantic mapping using an external PoE (Power over Ethernet) camera, calibrated and undistorted to be able to provide observations of the scene, including pose estimation and object localization.
- Transformation of image-space detections into real world coordinates through homography-based geometric mapping.
- Creation of an ontology, where prior knowledge can be stored, as well as enabling real time updates from sensory observations.
- Implementation and dynamic updating of semantic spatial relation that allows continuous reasoning about proximity relationships between a robot and other objects in the scene.

The organization of the paper is as follows: Section 2 describes the related work and consists of recent works in semantic mapping and representation of prior knowledge. Section 3 describes the methodology pipeline, including all its components in detail and an algorithm of the whole implementation. Finally, Section 4 concludes the paper and discusses future work.

## 2. Related work

Semantic mapping has emerged as an active research area in robotics with several frameworks and semantic mapping systems that have been proposed to support dynamic scene understanding and environment representation.

### 2.1. On semantic mapping

An online semantic mapping and navigation ROS architecture for ground exploration is presented in [16]. The architecture includes a 3D semantic mapping process, a 2.5D navigation graph builder and a semantic aware A* planner. The system integrates two semantic mapping methods, Semantic Octomap and Kimera Semantics. In the first one the mapping is performed based on the Octomap package which provides and octree-based voxel volumic mapping representation. The second one performs semantic mapping based on Voxblox, a TSDF surfacic mapping package based on voxel hashing and marching cubes. The navigation part includes an online 2.5D semantic navigation graph builder which collects data and integrates it into a structure for navigation and a weighted A* path finder which performs path planning inside the navigation graph layer. The system was evaluated in simulation built from a synthetic dataset and on an real-world dataset, as well as an autonomous exploration task with a robot. The main differences with our approach are the 3D semantic mapping, instead of our 2D, the integration of a path planner and the lack of any prior knowledge representation.

Cuaran et al [17] present an active semantic mapping approach in horticultural environments. They utilized a mobile robot manipulator equipped with an RGB-D camera and tried to create a semantic and geometric representation of plants. Their system included four modules, semantic extractor, mapping, viewpoint planner, planning and control. In the first, pixel labels corresponding to three semantic classes are extracted with semantic segmentation. For mapping, Semantic Octomap was used to obtain the map representation. The viewpoint planner refined an incomplete map by selecting the best viewpoints to improve reconstruction. Finally, in the planning and control module, the order of execution was computed. The system was evaluated in simulation and real world with a custom 6-DOF robotic arm on a wheeled platform. Here, the camera utilized is on the robot and even though there is some form of semantic and geometric representation these are not in the form of ontologies.

In [18], Achat et al present a study on semantic information for autonomous robot tasks with planning capabilities. In this proposal, data from the semantic map are integrated into a multi-layer structure and can be adapted to multiple planners. The semantic Octomap online 3D mapping algorithm was utilized for the map representation. In addition, modified versions of A*, T-RRT and a shortcut procedure were adapted. As the base case study, an exploration task with an autonomous robot has been studied. The study included the computation of high-level goals and path planning derived from semantic information. This study, again, utilizes 3D mapping and gives more emphasis to planning.

IntelliMove [19] is a semantic mapping framework for autonomous robotic navigation and decision making and it consists of two core modules. The first is a hierarchical semantic topometric map framework, named IntelliMap, in which metric semantic and topological semantic data are combined within a multilayer structure. Firstly, sensory data is collected and SLAM is used for the map production. In addition, an object mapping component utilizes this data to classify and organize objects spatially within the map. The second core module, semantic planning, uses the maps created by the first module to allow robots to develop and execute navigation and task strategies. The framework was evaluated in comparison with other semantic mapping methods and demonstrated the best results. This approach follows the typical semantic mapping pipeline with SLAM for the map creation, but it further improves it by adding the second module for navigation and task strategies.

A semantic mapping technique for scene-wise object localization and a technique for integrating scene-level information into large-scale mapping are proposed in [20]. A monocular camera is used to estimate distance and size of scene objects and bounding box coordinates are utilized for object localization in 2D images. The camera is mounted on a robot and the distances of the objects from that robot are represented with a 2D map. Information about the scene that was based on the 2D map is also utilized along with wheel odometry to create a large-scale map. In addition, the semantic mapping method can be integrated into a platform with limited-resources hardware. This is a work similar to ours. We both use a monocular camera for collecting data information as well as utilize bounding box coordinates of objects. Finally, both methods try to create 2D semantic maps.

### 2.2. On prior knowledge and graph representation

Xinghang et al [21] present a new framework for the embodied semantic scene graph generation problem, which leverages the physical capabilities of the intelligent agent to autonomously generate an appropriate path for environment exploration to be utilized for scene graph generation. The developed architecture includes the scene graph generation module and the navigation module. The local scene graph is generated from a 3D semantic point cloud, which is produced from RGB and depth images, while for navigation the next action and step is generated from the RGB frame, the previous action and the scene graph. In addition, imitation learning an reinforcement learning were utilized for pre-training and fine-tuning accordingly. The proposed method was tested and evaluated in a streaming video captioning task with promising results. The main difference with our approach is the use of scene graphs instead.

In [22], Amodeo et al present an ontology framework that can improve an existing machine learning based scene graph generator is introduced. The scene graph generator is augmented with ontology-based reasoning mechanism, as well as the dataset for the application with inferred knowledge. The proposed method includes two main processes, a training dataset filtering and augmentation process and a network output post-processing process. A scene graph generation network also needs to be available with object detection and processing capabilities, being able to receive dense semantic vectors with semantic information about the scene and finally, to output a single ranking value for every knowledge triplet proposal. For the training dataset filtering and augmentation step, a formal ontology is created and different ontology based transformations are applied to the original dataset. The output post-processing process prunes relations triplets that introduce violations in the previously created ontology. Multiple types of experiments were conducted, with different datasets and scene graph generations models, showed quantitative and qualitative improvements in the generated scene graphs. The ontology-based reasoning mechanism that is described here is similar to our approach.

The authors of [23] introduce a new graph representation that captures metric and semantic aspects of a dynamic environment, named 3D Dynamic Scene Graph (DSG). A DSG is a layered directed graph which consists of nodes representing spatial concepts and edges representing relations, while layers correspond to different levels of abstraction in the scene. They also present Kimera, a fully automatic spatial perception engine for building a DSG from visual data. Kimera consists of two modules, the first is responsible for real time metric-semantic reconstruction of the scene and includes four distinct submodules. A visual inertial odometry module, a local 3D mesh for collision avoidance module, a module that builds a global 3D mesh and annotates it, and a module that applies visual loop closure through the joint optimization of the robot trajectory pose graph and the global 3D mesh. The second module is responsible for building the DSG and includes three submodules. A submodule that reconstructs dense meshes of humans and calculates their trajectories, a submodule that computes a bounding boxes for objects that have unknown shape while fitting CAD models to objects with known shapes and the final one transforms the metric-semantic mesh into a topological graph of obstacle-free locations, performs room segmentation and identifies enclosing structures. The method was evaluated in real life datasets and photo realistic simulations and the results showed competitive performance in visual SLAM, accurate estimation of 3D metric semantic mesh model and the creation of a DSG for a complex indoor environment. Again, this is another work that utilizes scene graphs instead of ontologies for prior knowledge representation.

Dimitropoulos and Hatzilygeroudis [24] propose a combined approach for context representation, including a robotics-related ontology and its transformation to a knowledge graph. The ontology was created to be utilized for various robotic tasks and includes a plethora of classes with subclasses representing objects in a robotic lab or robot actions, as well as their properties. The ontology was evaluated with the use of the editor built in reasoner and outside of the editor with an online evaluation tool to be as consistent as possible. After that the ontology was converted to a knowledge graph, which then was queried with different types of queries, created in an intuitive and user-friendly graph query language. This is the work in which our ontology was based on. Even though our ontology was created from scratch it was heavily influenced by the ontology that is described in [24].

## 3. Material and Methods

Our goal was to represent the contextual knowledge of a robot movement environment, more specifically to create a map of a region inside a computer science laboratory and dynamically update it. We created an ontology to represent the knowledge inside the laboratory and utilized an external PoE (Power over Ethernet) camera to capture the scene.

YOLOv11s was trained to detect different kinds of objects from the camera stream. The camera was calibrated, undistorted and its sensory data could provide object localization and pose estimation of a NAO robot in the scene. In addition, homography transformation was utilized to find the real-world object coordinates. The ontology is dynamically updated every time the location of an object is changed, an object is added or removed from the scene, or when the robot got close enough to another object.

### 3.1. The ontology

We decided to use an ontology for knowledge representation for two reasons. First, the knowledge is represented in a hierarchical way that is easily readable and understandable. Second, it also provides easy access to python scripts to modify it and update it. Protégé was decided as the ontology editor. The ontology features 35 classes of entities, two of them being the main subclasses, “physical_entity” and “abstract_entity” (see Fig. 1 for a part of the ontology). Physical entities are mainly different laboratory objects, and the abstract entities represent observations. An *observation* consists of the pose (X, Y and theta for the robot), the confidence factor (*cf*) and the date time (*dt*) when a physical entity was detected in the scene:

$$obs = [X, Y, \theta, cf, dt]. \quad (1)$$

The concepts of X, Y, theta, confidence and date time were also represented as data properties with “observation” as domain. However, the above properties are not updated. They are just added every time a new observation occurs to keep a history of what has been happening as the time passes.

In addition, even though these data properties are connected to an object semantically, in reality their domain suggests that they are connected to a specific observation that happened in a specific time frame. For these reasons, other similar data properties were also added in the ontology that represent the current coordinates that an object has and the confidence factor and the last time when that object was observed at. These data properties are updated dynamically and have the "physical_entity" as domain. Finally, the ontology includes two object properties. The first one, "observes", has "observation" as domain and "physical entity" as range and accompanies the first set of data properties. The second, "is_near_to", is symmetric and is updated when the robot is close enough to another object. There are no individuals created by hand, they are all being added, removed and updated dynamically (automatically).

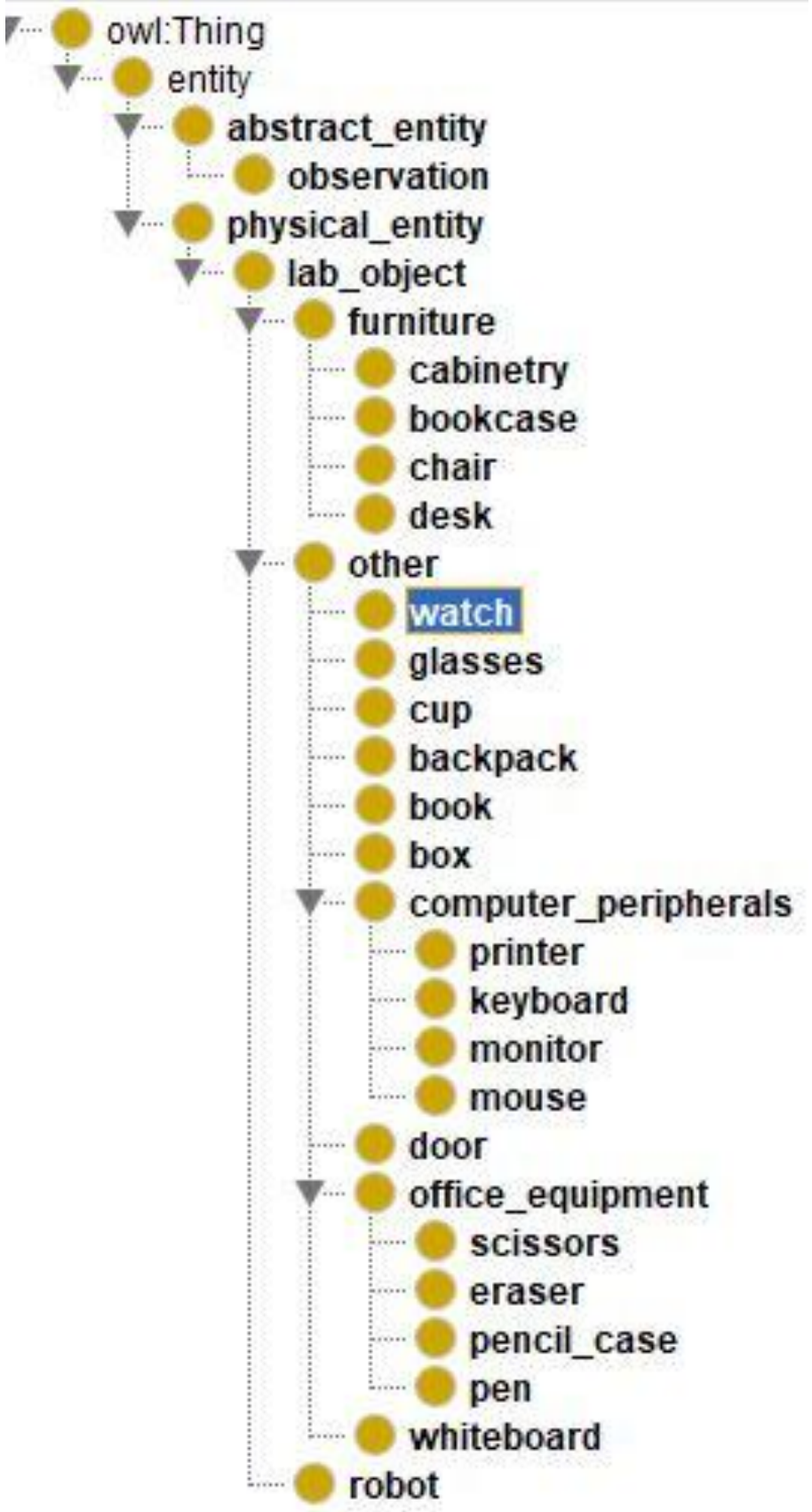

***Fig.1*** *Part of the ontology*

## 3.2. The camera and its Calibration

### *3.2.1 Camera identity*

Tapo C501GW is the external camera that we utilized in our application. It is a PoE/IP camera that supports connection through ethernet for the lowest latency possible, stable bandwidth and no packet loss. It's also a Full HD camera with 360° horizontal and 130° vertical viewing angles. We put the camera high on the wall, at around 3-3.5 meters, so that it can cover the whole scene. The camera also supports RTSP (Real-Time Streaming Protocol) for streaming and we used that for NAO localization, object detection and converting pixel coordinates to real ones. We experimented with the idea of using the robot's camera, but it wasn't good enough for our purpose mainly due to its hardware limitations.

### *3.2.2 Fixing distortion*

In order to find the real-world coordinates of our objects, a homography transformation was considered. However, because of the camera's wide angle lenses the straight lines in the test region of the lab appeared distorted in the captured images. This is very important because it can cause errors during the process of the transformation of the measured coordinates to real world ones, and inaccuracy in navigation. In order to correct the distortion, we calibrated our camera with OpenCV. During calibration the intrinsic camera parameters and distortion coefficients are computed. The camera matrix $K$ describes how the 3D camera coordinates system are projected onto the two-dimensional image plane:

$$K = \begin{bmatrix} f_x & 0 & c_x \\ 0 & f_y & c_y \\ 0 & 0 & 1 \end{bmatrix} \tag{2}$$

where: $f_x, f_y$:focal lengths and $c_x, c_y$: principal point.

Distortion is described through distortion coefficients:

$$(k_1, k_2, p_1, p_2, k_3)$$

where:

- $k_1, k_2, k_3$: radial distortion,
- $p_1, p_2$: tangential distortion.

We estimated the distortion coefficients using a checkerboard pattern. The pattern was of 9x6 size and multiple images of it were captured in different angles and different locations inside the test area, while remaining visual to the camera. The detected checkerboard corners were then used by the OpenCV calibration algorithm to compute the above camera matrix and distortion coefficients. Both the original $K$ and the distortion coefficients are used as arguments in the OpenCV function: *getOptimalNewCameraMatrix().* This function computes the new optimized projection matrix $K$':

$$K' = \begin{bmatrix} f'_x & 0 & c'_x \\ 0 & f'_y & c'_y \\ 0 & 0 & 1 \end{bmatrix} \tag{3}$$

where: $f'_x, f'_y$: the new focal lengths and $c'_x, c'_y$: the new principal points

After that, $K'$ is used as argument in another function: *undistort()* which undistorts the original image.

With this way, the camera was calibrated, and the calibration settings were stored to be utilized to undistort each incoming video frame. So, every frame in the live perception pipeline is being undistorted first and appeared mostly geometrically correct.

*3.2.3 Real world coordinates estimation*

Our test region was approximately of 180x240cm size. This region constitutes our real world coordinate system. However, because the images taken from the camera covered a bigger area than desired, we defined our test region by selecting the four corners in a picture that represented our four real world determination points. The points in the picture correspond approximately to the following real world points:

$$[0,0], [180,0][180,240], [0,240]$$

With these world points the projective transformation homography matrix *H* was calculated:

$$H = \begin{bmatrix} h_{11} & h_{12} & h_{13} \\ h_{21} & h_{22} & h_{23} \\ h_{31} & h_{32} & h_{33} \end{bmatrix} \tag{4}$$

and

$$\lambda \begin{bmatrix} x \\ y \\ 1 \end{bmatrix} = H \begin{bmatrix} u \\ v \\ 1 \end{bmatrix} \tag{5}$$

where

- $(u, v)$: image coordinates
- $(x, y)$: world coordinates
- $\lambda$ : scale factor.

The nine coefficients $h_{ij}$ in matrix *H* define the projective transformation between image coordinates and world coordinates. These coefficients are computed from the four selected point pairs described above.

$\lambda$ is a projective scale factor, which is introduced because homography operates in homogenous coordinates. After multiplication of the image coordinates by the homography matrix, the result is normalized by dividing it by $\lambda$ and this is how the final real-world coordinates are estimated.

After computation of the homography matrix *H* the homography calibration was saved. During our system run time every detected object or marker was transformed from pixel coordinates to world coordinates (cm).

Fig.2 and Fig.3 show a distorted and undistorted image captured by the camera as well as the four real world points represented by the blue circles in the undistorted image.

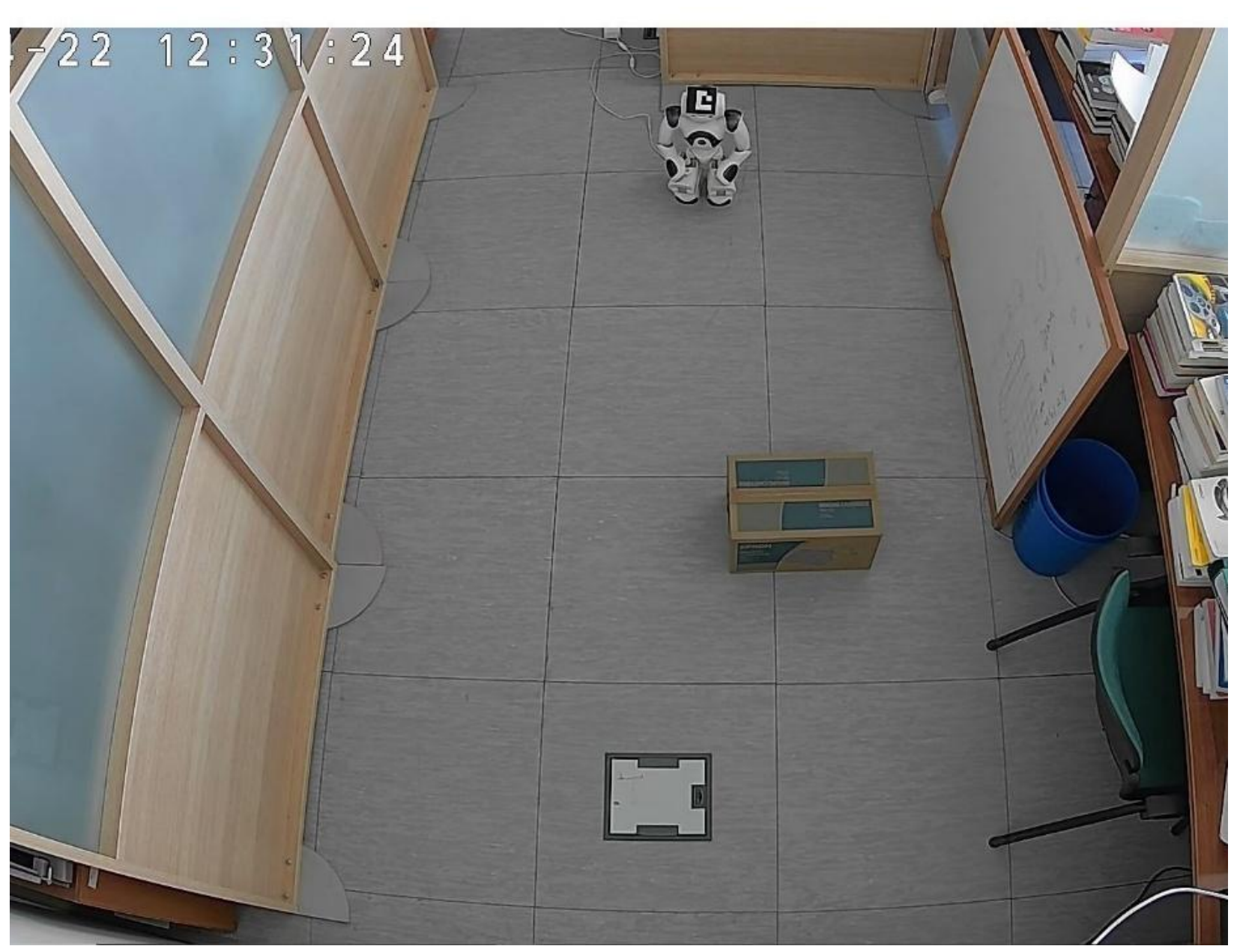

***Fig.2*** *Distorted image*

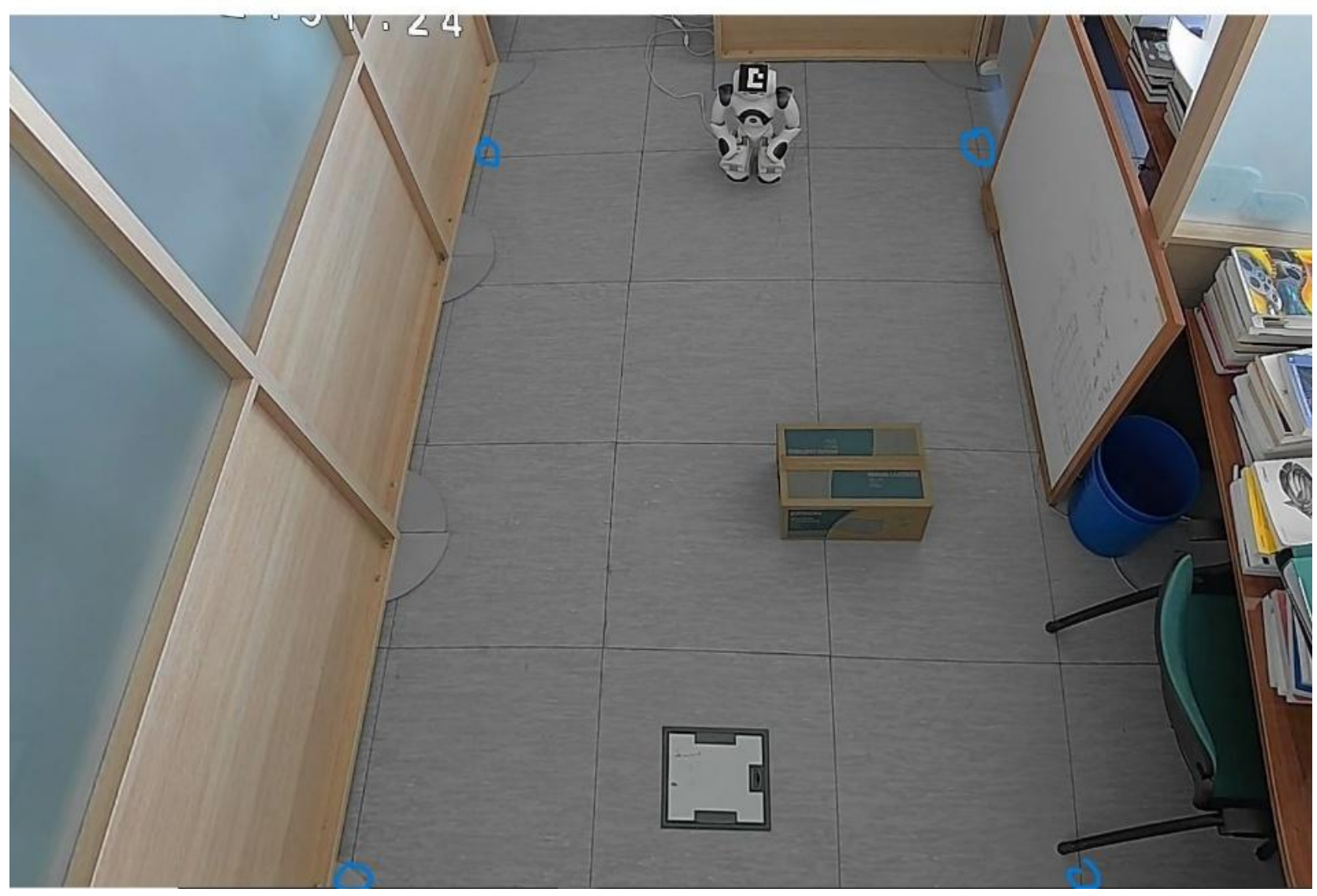

***Fig.3*** *Undistorted image*

*3.2.4 Camera calibration process*

Our camera calibration process is represented by the flowchart of Fig. 4. First, the image captured by the camera appears distorted because of the camera's wide lenses. In order to undistort it, images of a checkerboard pattern were captured in the test region. The camera matrix and the distortion coefficients were computed based on those captured images. After that every frame in the camera feed appeared undistorted. We defined our test region by selecting four points in an undistorted image captured by our camera, that represented four real world points. After that, the transformation homography matrix was estimated based on those points and the calibration was saved. Finally, during system run time pixel coordinates of objects detected in an image were transformed into real world coordinates.

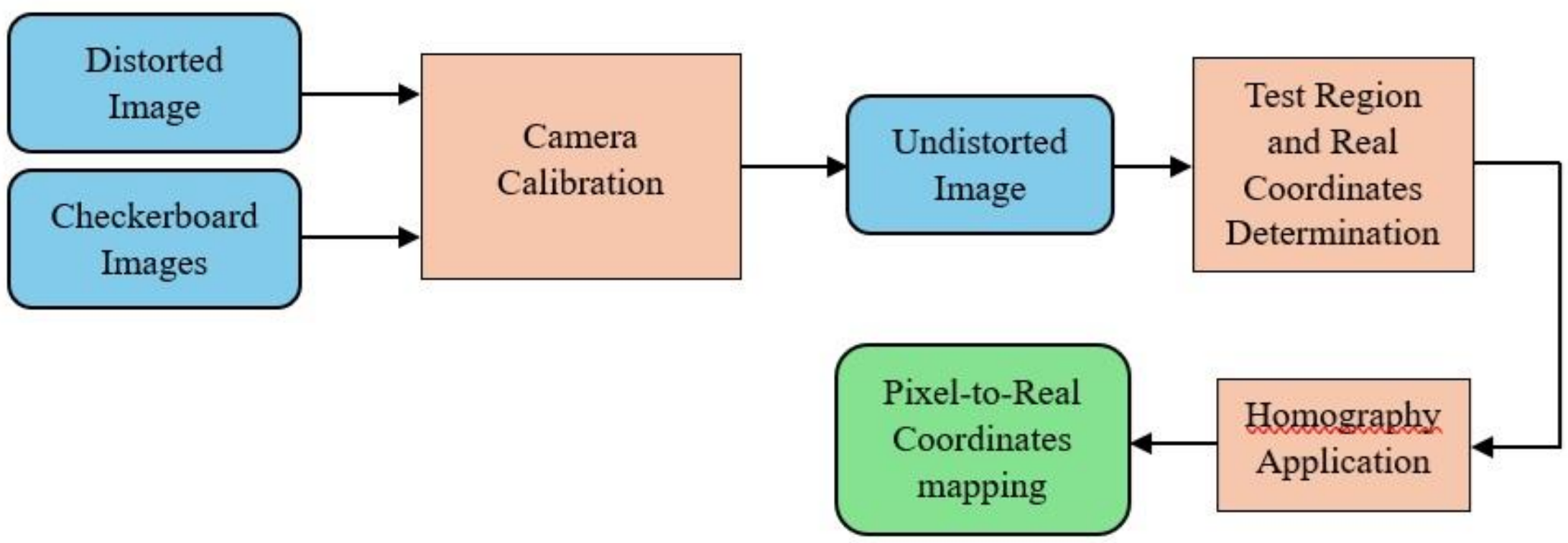

***Fig.4*** *Camera calibration process flowchart*

### 3.3. Robot localization

A stable robot detection and accurate pose estimation is important for navigation. While an object detection model could detect the robot, it could not provide an accurate enough position and orientation information. This was the main reason that we decided to utilize an ArUco marker. Generally, the ArUco marker provides a more stable detection with less jitter than a typical YOLO bounding box and it is also more durable to small lightning changes. This stability is very important for updating the ontology and future navigation tasks. Finally, because it is deterministic and unique, it doesn't require any training and provides unambiguous identification. The marker was put on the head of the robot in horizontal position (see Fig. 5). First, the center of the detected marker was computed as the average of its four corners coordinates:

$$c_x = \frac{1}{4}\sum_{i=1}^{4} x_i \quad (6)$$

$$c_y = \frac{1}{4}\sum_{i=1}^{4} y_i \quad (7)$$

Where $(x_i, y_i), i = 1...4$ are the four corners of the marker.

Finally, the homography transformation was applied to find the resulting point.

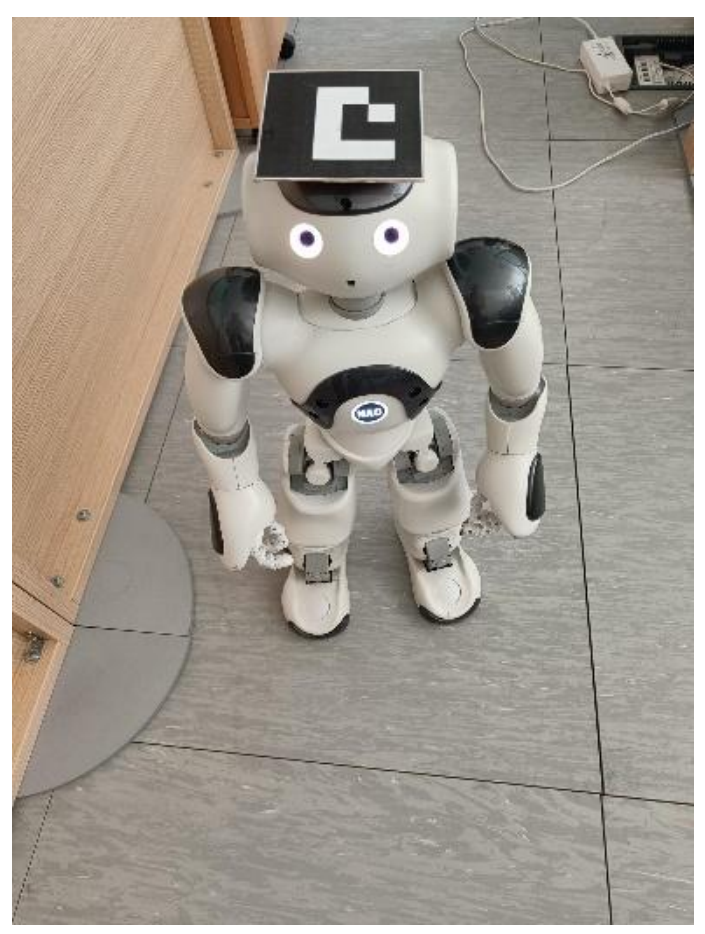

***Fig.5*** *NAO robot and the marker on its head*

### 3.4. Object detection and real world coordinates estimation

To get as accurate observations for our ontology as possible YOLOv11s was utilized for object detection. We wanted the model to be able to recognize 10 classes of objects that could fit inside our test region.

In the beginning, we tried a lighter model, YOLOv4-tiny, to get the results as fast as possible. However, because of its tiny version and smaller backbone, the model had trouble detecting smaller objects. In fact, the only objects that were consistently detected were larger objects like chairs or big boxes and everything else, especially objects that were not close to the camera, were not detected at all. In addition, because we not only wanted a model that could detect objects, but also one that would feed the ontology with observations and update it accordingly, the detection should be very accurate. So, accuracy was our main concern but being in a good balance with response time. With this in mind, we concluded to YOLOv11s, one of the newest models of the YOLO family, which is very accurate and because of its small edition, speedy enough too [25].

We trained the model mainly with 4 datasets, Open Images dataset V7 from Google, 2 datasets from Raboflow and a custom one with images captured by our external camera because we wanted to cover the viewpoint of the camera in the application. We used 5000 images from the three and captured 400 more with our camera. We validated the training with 40 images, captured again by our camera with multiple objects directly in the scene. A mAP of 99.5% was achieved with over 90% confidence factor for the 10 classes of objects: monitor, keyboard, book, bottle, scissors, cup, cabinetry, backpack, box, chair.

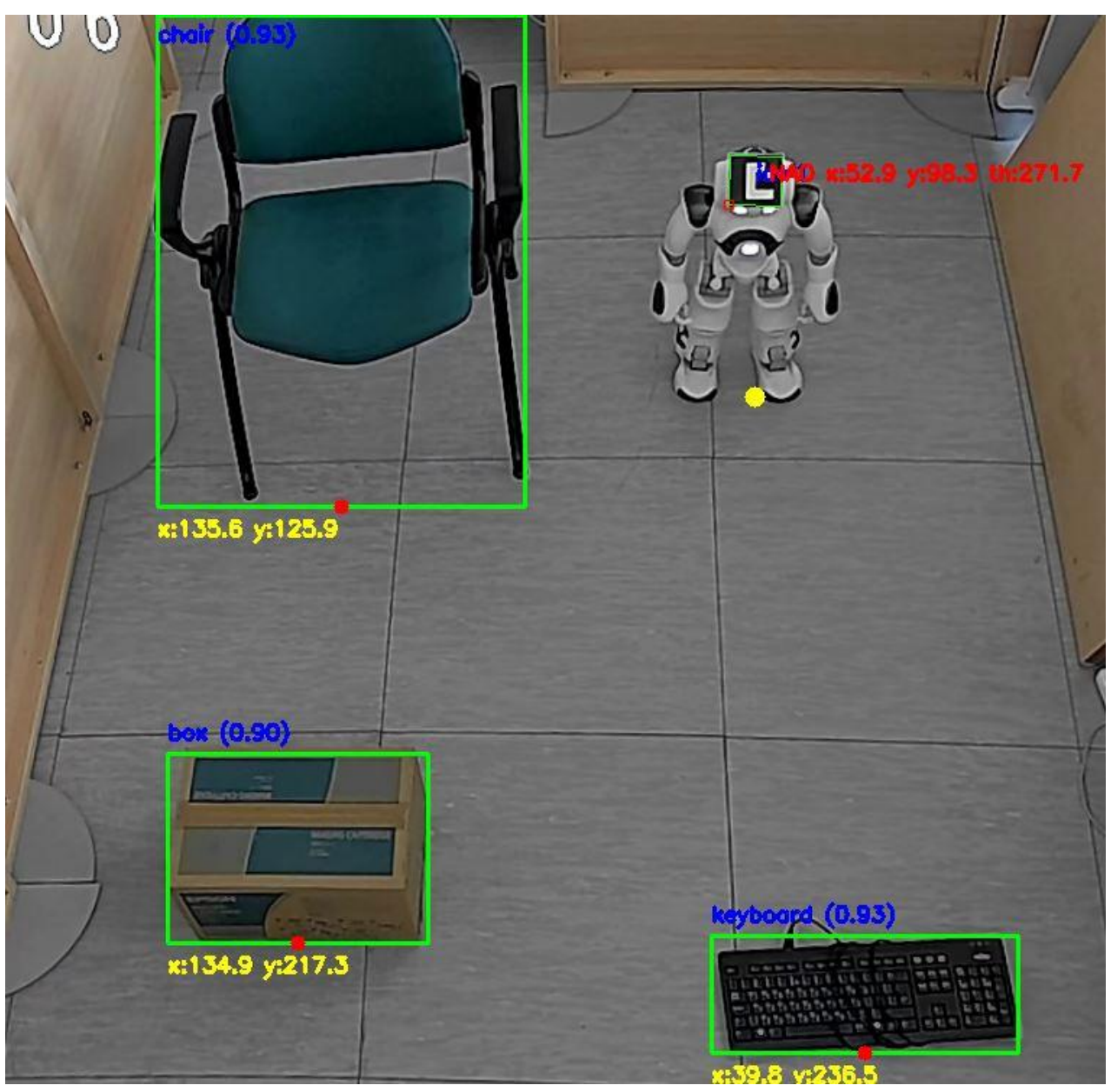

***Fig.6*** *Object detection.*

Given the coordinates of the upper left and bottom right corner, ($x_1, y_1$) and ($x_2, y_2$) respectively, the coordinates of the middle of the bottom side of the bounding box, which is considered as the physical contact point, are calculated as follows:

$$c_x = \frac{x_1+x_2}{2} \tag{8}$$

$$c_y = y_2. \tag{9}$$

Each object in the scene touches the ground, so this is the correct way of computing the location of the objects, because this is the actual point of contact with the ground. After that we apply the homography transformation and obtain the real world coordinates: *x_world* and *y_world* of each object.

### 3.5. Correcting estimated real world coordinates

To check our methodology for estimating real world coordinates, we conducted a number of experiments regarding real world coordinates estimation for NAO robot and various objects in our lab. During the experiments, the robot and eight different objects were put in different positions inside the test area while the system was running so that their coordinates could be estimated online. We present the results below for the objects (Table 1) and the NAO robot (Table 2). Obviously, $x \in \{0,180\}$ and $y \in \{0,240\}$ .

As shown in Table1, the average localization error for the objects is 5.3cm along the x-axis and 17.8cm along the y-axis. This can be attributed to the oblique camera viewpoint, the perspective compression along the depth direction and the residual lens distortion even though the camera was calibrated. Respectively (see Table 2), for the robot, the average localization error along the x-axis is 17.7cm and even greater (28.1cm) along y-axis, which are quite big. This is probably due to the fact that ArUco marker is located on the head of the robot and does not have any contact with the ground.

**Table 1.** Real vs Estimated real world objects' coordinates

| Object | Real x (cm) | Real y (cm) | Estimated x (cm) | Estimated y (cm) | \|Diff(x)\| | \|Diff(y)\| |
|---|---|---|---|---|---|---|
| **Monitor** | **58** | **96** | 60 | 128 | 2 | 32 |
| **Monitor** | **131** | **47** | 142 | 68 | 11 | 21 |
| **Keyboard** | **137** | **166** | 138 | 186 | 1 | 22 |
| **Keyboard** | **40** | **176** | 38 | 182 | 2 | 6 |
| **Box** | **74** | **174** | 79 | 193 | 5 | 19 |
| **Box** | **35** | **135** | 32 | 147 | 3 | 12 |
| **Book** | **45** | **54** | 50 | 93 | 5 | 39 |
| **Book** | **34** | **218** | 32 | 219 | 2 | 1 |
| **Cup** | **140** | **152** | 145 | 173 | 5 | 21 |
| **Cup** | **143** | **203** | 148 | 209 | 5 | 6 |
| **Chair** | **37** | **223** | 33 | 237 | 4 | 14 |
| **Chair** | **30** | **79** | 21 | 97 | 9 | 18 |
| **Bottle** | **92** | **125** | 98 | 152 | 6 | 27 |
| **Bottle** | **89** | **192** | 92 | 194 | 3 | 2 |
| **Cabinetry** | **120** | **52** | 132 | 93 | 12 | 41 |
| **Cabinetry** | **135** | **168** | 146 | 173 | 11 | 5 |
| **MAE** | | | | | **5.3** | **17.8** |

**Table 2.** Real vs Estimated real world NAO coordinates

| Position | Real x (cm) | Real y (cm) | Estimated x (cm) | Estimated y (cm) | \|Diff(x)\| | \|Diff(y)\| |
|---|---|---|---|---|---|---|
| **1** | 87 | **47** | 96 | 9 | 9 | 36 |
| **2** | 146 | **45** | 172 | 11 | 26 | 34 |
| **3** | 136 | **114** | 160 | 83 | 24 | 31 |
| **4** | 84 | **155** | 92 | 125 | 8 | 30 |
| **5** | 34 | **129** | 20 | 99 | 14 | 30 |
| **6** | 25 | **236** | 8 | 219 | 17 | 17 |
| **7** | 162 | **227** | 188 | 208 | 26 | 19 |
| **MAE** | | | | | **17.7** | **28.1** |

Those results were not satisfactory for our purpose, given that we set as target limit to have a mean absolute error (MAE) less than 1% of the diagonal of the test area, i.e. 3cm, across both dimensions (x, y).

To achieve that, given that the sets of real and estimated values in all cases are highly linear (see Fig. 7 for x-axis values), we employed a linear regression (LR) model to correct coordinates produced by the calibration process. More specifically, we employed different models for correction of x-axis and y-axis coordinates for both the objects and the NAO robot. We employed different models for the NAO robot, given that position estimation for NAO is done in a different way than that of objects.

Based on the real and estimated x-axis data of Table 1, we developed the following LR model for objects' x-axis values (see Fig. 7 for its graphical representation):

$$x_{cor} = 5.352 + 0.905\, x_{est} \tag{10}$$

where $x_{cor}$ is the corrected value and $x_{est}$ the estimated value produced by the calibration process.

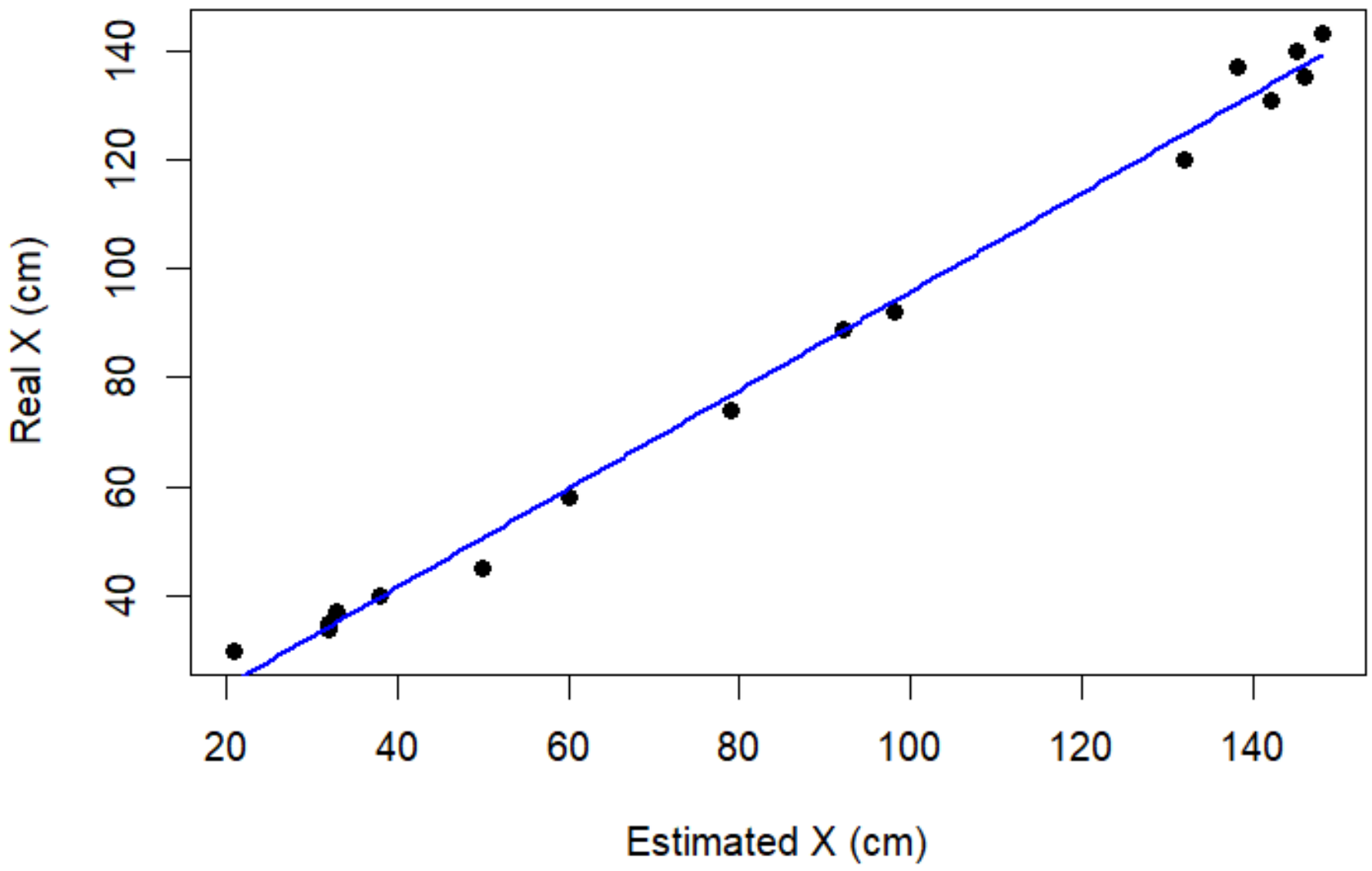

***Fig.*** *7 Linear Regression for x-axis data.*

The above LR model exhibited an excellent fit ($R^2$=0.9936, adjusted $R^2$=0.9932), indicating that approximately 99.4% of the variance in the real (corrected) coordinates was explained by the estimated coordinates. The regression coefficient was highly significant (p<0.000…1). The model achieved a mean absolute error (MAE) of approximately 2.8 cm. Leave-one-out cross-validation produced prediction errors comparable to those obtained on the calibration data, demonstrating good generalization performance. This is a confirmation of the adequacy of the sample size.

**Table 3.** Real vs Corrected objects' coordinates

| Object | Real x (cm) | Real y (cm) | Corrected x (cm) | Corrected y (cm) | \|Diff(x)\| | \|Diff(y)\| |
|---|---|---|---|---|---|---|
| **Monitor** | **58** | **96** | 60 | 105 | 2 | 9 |
| **Monitor** | **131** | **47** | 134 | 35 | 3 | 12 |
| **Keyboard** | **137** | **166** | 130 | 173 | 7 | 7 |
| **Keyboard** | **40** | **176** | 40 | 168 | 0 | 8 |
| **Box** | **74** | **174** | 77 | 181 | 3 | 7 |
| **Box** | **35** | **135** | 34 | 127 | 1 | 8 |
| **Book** | **45** | **54** | 51 | 64 | 6 | 10 |
| **Book** | **34** | **218** | 34 | 211 | 0 | 7 |
| **Cup** | **140** | **152** | 136 | 158 | 4 | 6 |
| **Cup** | **143** | **203** | 139 | 200 | 4 | 3 |
| **Chair** | **37** | **223** | 35 | 232 | 2 | 9 |
| **Chair** | **30** | **79** | 24 | 69 | 6 | 10 |
| **Bottle** | **92** | **125** | 94 | 133 | 2 | 8 |
| **Bottle** | **89** | **192** | 89 | 182 | 0 | 10 |
| **Cabinetry** | **120** | **52** | 12 | 64 | 5 | 12 |
| **Cabinetry** | **135** | **168** | 137 | 158 | 2 | 10 |
| **MAE** | | | | | **2.7** | **8.5** |

In the same way, the following LR model was developed for the y-axis case:

$$y_{cor} = -44.476 + 1.168\, y_{est} \tag{11}$$

with very similar statistics and a MAE of 8.5cm.

In Table 3, the corrected values of the coordinates of the objects are presented in comparison to the real ones.

Accordingly, the following LR models were developed for the NAO robot case:

$$x_{r_cor} = 17.574 + 0.749\, x_{r_est} \quad (12)$$

$$y_{r_cor} = 37.930 + 0.912\, y_{r_est} \quad (13)$$

achieving a MAE of 1.6cm in both cases.

The above LR models also exhibited an excellent fit (e.g. $R^2$=0.9982, adjusted $R^2$=0.9978). This indicats that approximately 99.8% of the variance in the real (corrected) coordinates was explained by the estimated coordinates. The regression coefficients were highly significant (p<0.000…1). The models achieved a mean absolute error (MAE) of approximately 1.6 cm. Leave-one-out cross-validation produced prediction errors comparable to those obtained on the calibration data, demonstrating good generalization performance. This is a confirmation of the adequacy of this sample size too.

In Table 4, the corrected values of the coordinates of the NAO robot are presented in comparison to the real ones.

**Table 4.** Real vs Corrected real world NAO coordinates

| **Position** | **Real x (cm)** | **Real y (cm)** | **Corrected x (cm)** | **Corrected y (cm)** | **\|Diff(x)\|** | **\|Diff(y)\|** |
|---|---|---|---|---|---|---|
| **1** | 87 | **47** | 89 | 46 | 2 | 1 |
| **2** | 146 | **45** | 146 | 48 | 0 | 3 |
| **3** | 136 | **114** | 137 | 114 | 1 | 0 |
| **4** | 84 | **155** | 86 | 152 | 2 | 3 |
| **5** | 34 | **129** | 33 | 128 | 1 | 1 |
| **6** | 25 | **236** | 24 | 238 | 1 | 2 |
| **7** | 162 | **227** | 158 | 228 | 4 | 1 |
| **MAE** | | | | | **1.57** | **1.57** |

**Table 5.** Real vs Corrected objects' coordinates on test set

| **Object** | **Real x (cm)** | **Real y (cm)** | **Corrected x (cm)** | **Corrected y (cm)** | **\|Diff(x)\|** | **\|Diff(y)\|** |
|---|---|---|---|---|---|---|
| **Bottle** | **150** | **150** | 150 | 147 | 0 | 3 |
| **Book** | **61** | **74** | 64 | 73 | 3 | 1 |
| **Keyboard** | **150** | **220** | 147 | 216 | 3 | 4 |
| **Cup** | **14** | **19** | 16 | 23 | 2 | 4 |
| **Cabinetry** | **88** | **161** | 91 | 155 | 3 | 6 |
| **Monitor** | **28** | **116** | 24 | 115 | 4 | 1 |
| **Chair** | **131** | **81** | 138 | 82 | 7 | 1 |
| **Backpack** | **150** | **220** | 147 | 216 | 3 | 4 |
| **Scissors** | **32** | **217** | 34 | 215 | 2 | 2 |
| **Box** | **30** | **167** | 27 | 166 | 3 | 1 |
| **MAE** | | | | | **2.7** | **2.4** |

**Table 6.** Real vs Corrected real world NAO coordinates on test set

| Position | Real x (cm) | Real y (cm) | Corrected x (cm) | Corrected y (cm) | \|Diff(x)\| | \|Diff(y)\| |
|---|---|---|---|---|---|---|
| **1** | **146** | **44** | 146 | 45 | 0 | 1 |
| **2** | **95** | **83** | 97 | 82 | 2 | 1 |
| **3** | **26** | **90** | 26 | 89 | 0 | 1 |
| **4** | **60** | **144** | 61 | 141 | 1 | 3 |
| **5** | **115** | **163** | 115 | 162 | 0 | 1 |
| **6** | **145** | **215** | 143 | 216 | 2 | 1 |
| **7** | **91** | **238** | 90 | 239 | 1 | 1 |
| **MAE** | | | | | **0.9** | **1.3** |

To test our models on non-seen data, we produced new instances using the above developed LR models. Furthermore, we included instances of objects not existing in the initial data sets, like scissors and backpack (see Fig. 8). The results are presented in Tables 5 and 6 and show even better accuracy, further reducing the MAE. Notice that in case of y-axis data for objects (Table 5) it reduced to 2.4cm, even less than the one for x-axis data. From Tables 5 and 6, it is obvious that all models keep the constraint of having a MAE < 3cm.

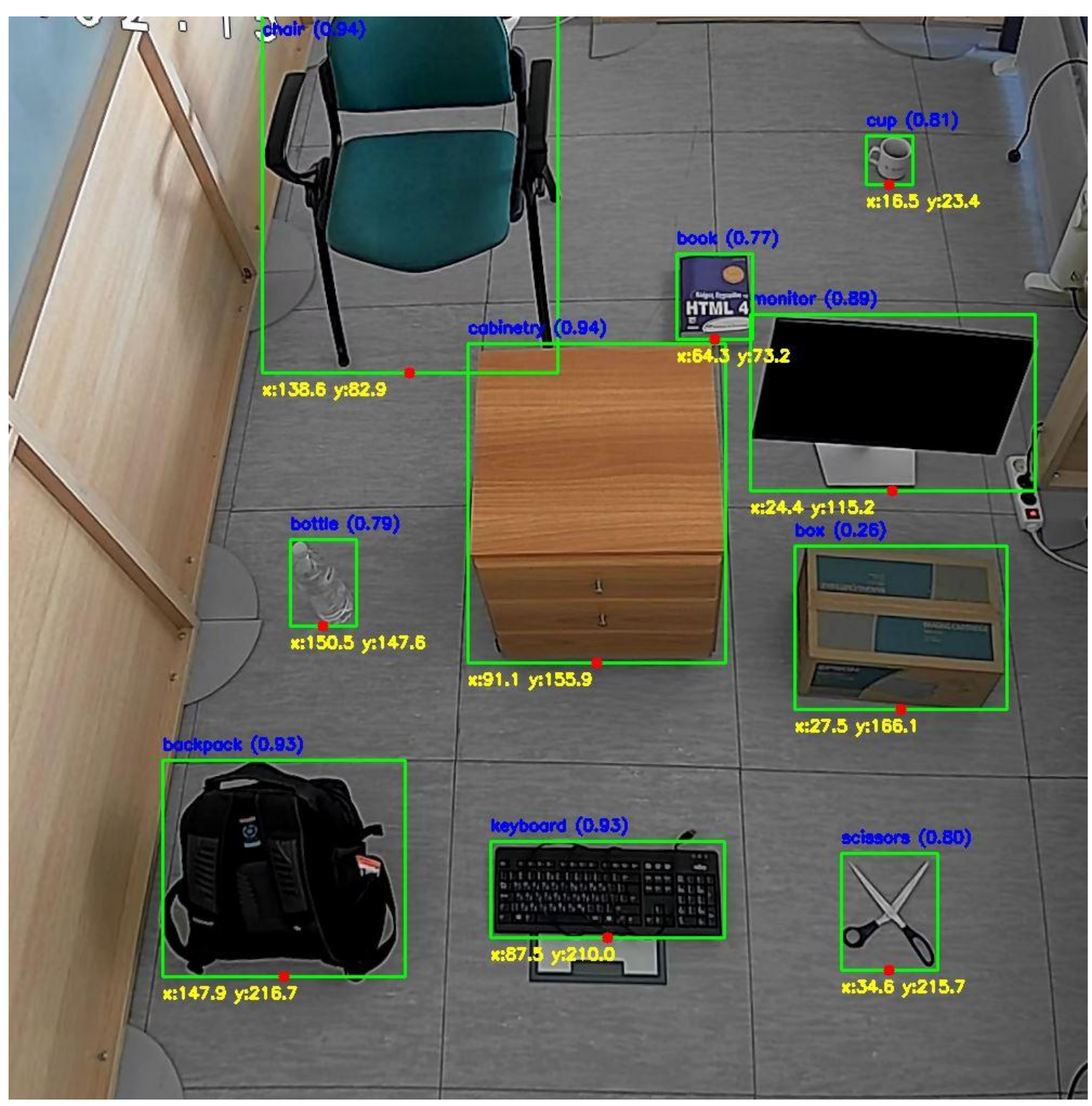

***Fig.8*** *Object Recognition with YOLOv11s.*

### 3.6. Updating the ontology

The ontology is updated dynamically every 5 seconds with new observations of the objects in the live stream. However, the data properties that represent the current coordinates of the objects are added dynamically only with the first observation and then they are updated only when their position is changed over a defined threshold. We introduce this threshold (~ 5cm) so that small fluctuations in the detection can be disregarded. If there are more than one object of the same class in the scene (see Fig. 6), all of them are added as different individuals of the same class. However, because the model cannot distinguish whether an object was removed from the scene and another one similar to that one just added, no objects would be removed or added in the ontology, only the coordinates of that object would be updated if its position is changed over the specific threshold. We try to find which object's position is changed using nearest-neighbor matching. If an observation indicates more new objects, these are added to the ontology. Now, if the observation shows less objects than those the ontology, the missing object is not removed immediately but only after it does not appear again in the next few detections. This is because the object may have not being detected in a frame but being detected normally in the next ones.

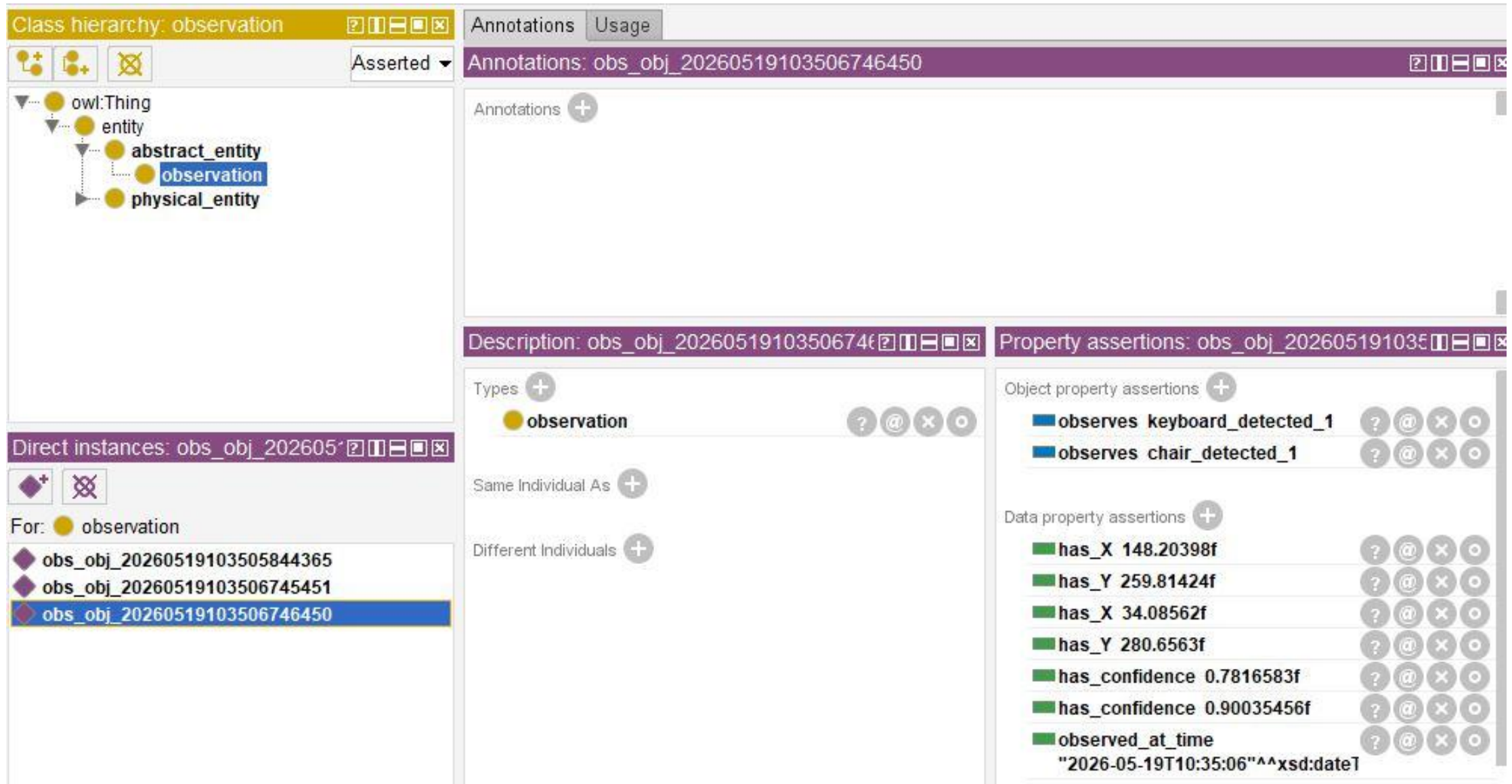

*Fig.7 Observations*

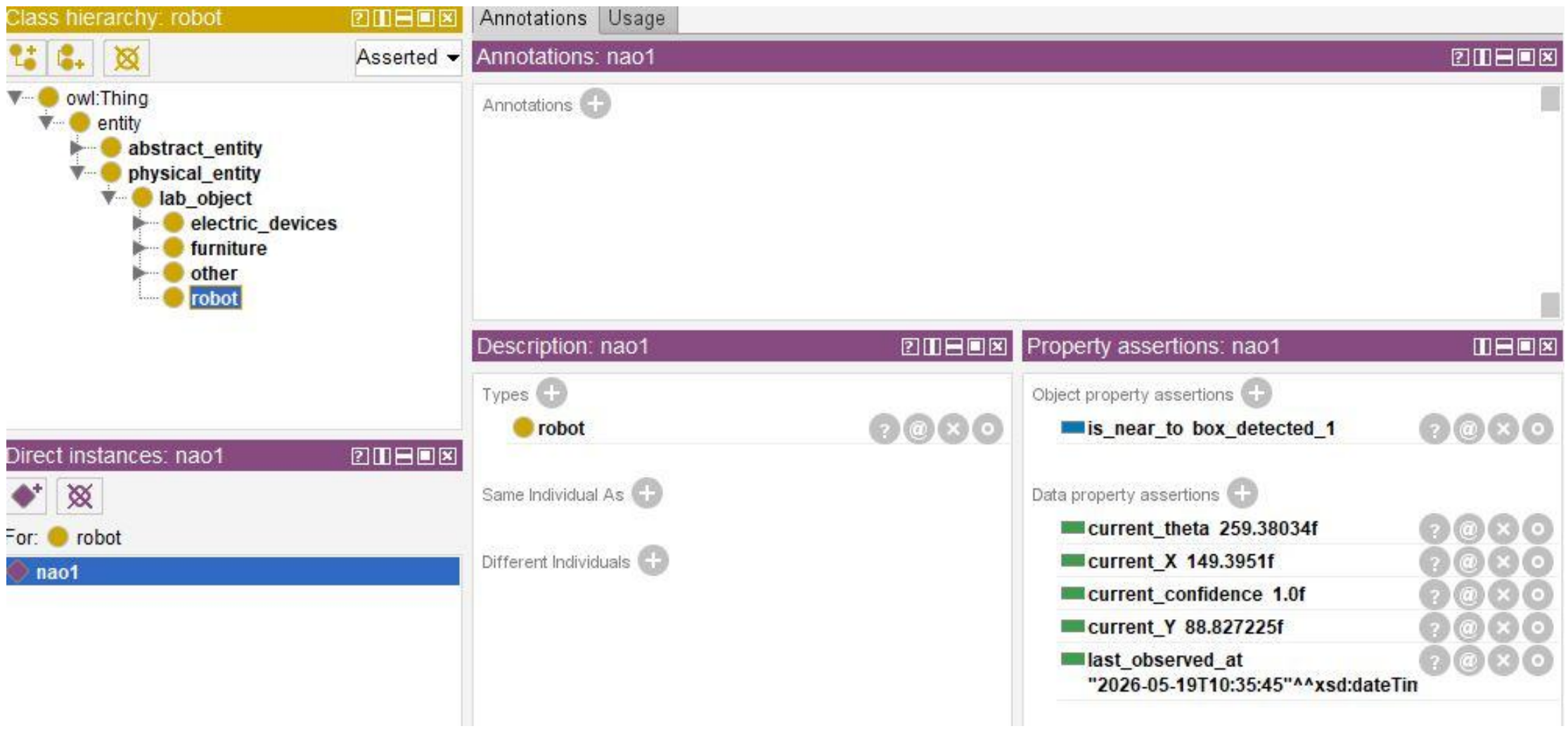

***Fig.8*** *"is_near_to" property*

The “is_near_to” object property is being updated every time the robot is near to an object (see Fig. 8). This property is calculated with the Euclidean distance of the “current_X”/“current_Y” data property of the robot and the object. There is another threshold (10cm) to determine whether the distance is considered “near” or not. Again, if previous observations are not currently hold true (i.e. corresponding objects are missing), they are removed, and the property is updated accordingly. Removal is approved after situation remains for three consecutive frames. The object property “is_near_to” refers to a task-level proximity.

### 3.7. Running System

The algorithm of the running system is presented below and the corresponding flowchart in Fig. 9.

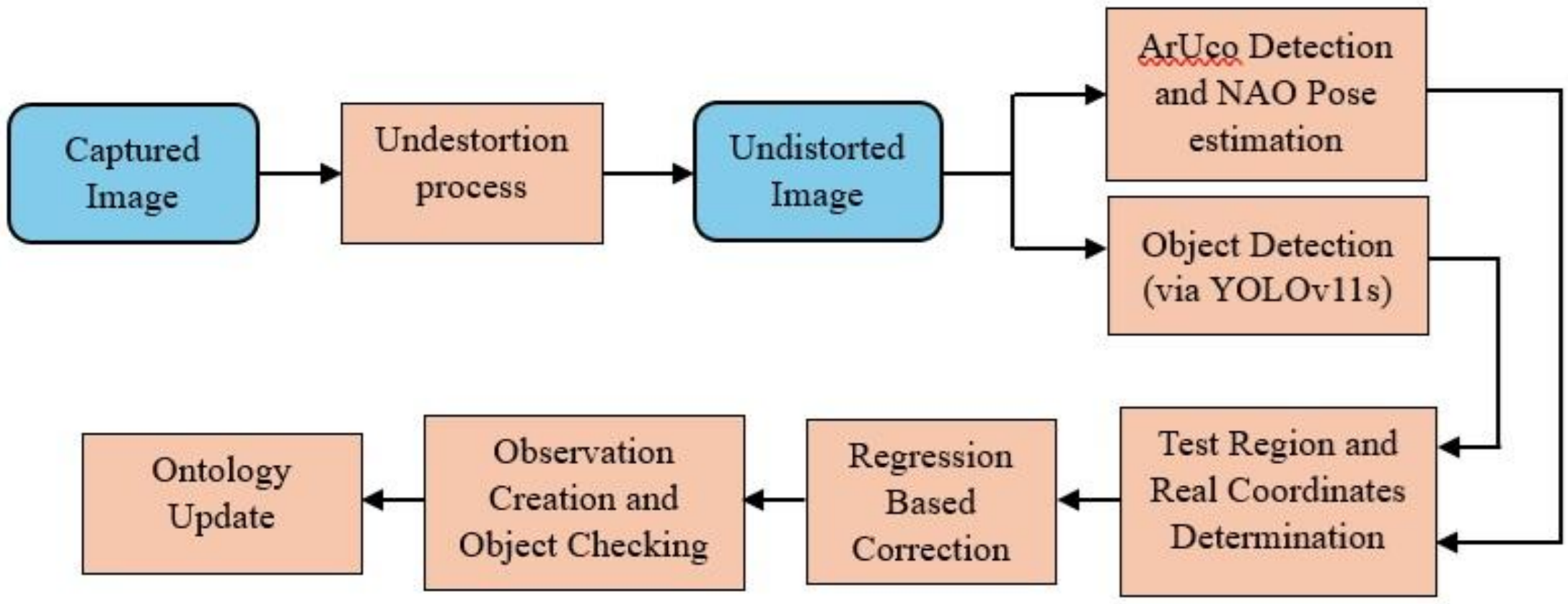

***Fig.9*** *The hybrid pipeline for dynamic semantic mapping*

**<u>Algorithm for running system</u>**

```
while RTSP stream is active do
       acquire frame
       undistort (frame)
       detect ArUco marker and estimate robot position
       run YOLO V11s on frame
       for each object detection do
              transform image coordinates to real world coordinates
              correct produced coordinates using regression models
              create ontology observation
              check objects and update ontology
       end for
       activate frame delay
       cleanup lost objects
       compute is_near_to object property
       save ontology
end while
```

First, the frame is acquired from the RTSP camera stream. Then the frame becomes undistorted using already calculated distortion coefficients. The robot pose (X, Y, θ) is estimated by detecting the ArUco marker on its head. After that the object detection model runs on that frame. For each detection the image coordinates are transformed into real world and new observations are added in the ontology. The ontology is updated accordingly (adding, updating individuals). If there have been enough frames with less detected objects, they are removed from the ontology and then the "is_near_to" property is computed. Finally, the ontology is saved.

## 4. Conclusion

In this paper a hybrid approach for dynamic ontology-based semantic mapping is presented. We designed and implemented a system with an external calibrated camera that utilizes projection homography for geometric mapping and localization. To that end, a linear regression-based correction of estimated values of real world coordinates is employed. In addition, the integration of YOLOv11s provides consistent object detection and tracking, while a developed ontology of entities in the scene stores prior knowledge and enables real time updates through observations from the cameras sensory data. Finally, the introduction and dynamic updating of the spatial relation "is_near_to" in the ontology enables continuous reasoning about proximity relationships between our robot and other objects in the scene.

Future work includes the integration of this ontology-based semantic mapping system into our robotic application that consists of task planning and seamless navigation inside an computer science lab.